\documentclass[sigconf]{acmart}
\pdfoutput=1
\renewcommand\footnotetextcopyrightpermission[1]{}
\usepackage{algorithm}
\usepackage{svg}
\usepackage{algpseudocode}
\usepackage{tabularx,booktabs}
\newcolumntype{Y}{>{\centering\arraybackslash}X}
\usepackage{amsmath}
\usepackage{graphicx}
\usepackage{caption}
\usepackage{subcaption}
\usepackage{float}
\setcopyright{none}
\acmConference[Proc. ICPR Workshop on VAIB]{Proc. ICPR Workshop on VAIB, January 2021}{January, 2021}{Milan, Italy}

\acmPrice{15.00}
\acmISBN{978-1-4503-XXXX-X/18/06}

\begin{document}
\settopmatter{printacmref=false} 

\title[Great Ape Facial Recognition]{A Dataset and Application for Facial Recognition of Individual Gorillas in Zoo Environments}


\author{Otto Brookes}
\affiliation{%
  \institution{University of Bristol, UK}
  \department{Dept. of Computer Science}}
\email{dl18206@bristol.ac.uk}

\author{Tilo Burghardt}
\affiliation{%
  \institution{University of Bristol, UK}
  \department{Dept. of Computer Science}}
\email{tilo@cs.bris.ac.uk}

\renewcommand{\shortauthors}{Brookes and Burghardt, et al.}

\begin{abstract}
We put forward a video dataset with 5k+~facial bounding box annotations across a troop of 7~western lowland gorillas~(\textit{Gorilla gorilla gorilla}) at Bristol Zoo Gardens. Training on this dataset, we implement and evaluate a standard deep learning pipeline on the task of facially recognising individual gorillas in a zoo environment. We show that a basic YOLOv3-powered application is able to perform identifications at 92\%~mAP when utilising single frames only. Tracking-by-detection-association and identity voting across short tracklets yields an improved robust performance at 97\%~mAP. To facilitate easy utilisation for enriching the research capabilities of zoo environments, we publish the code, video dataset, weights, and ground-truth annotations at data.bris.ac.uk.
\end{abstract}

\maketitle

\section{Introduction}
\textit{Motivation.} One important area of focus for zoos and sanctuaries is animal welfare~\cite{fraser2009assessing} and associated research questions: e.g. does captivity prohibit animals from functioning in a beneficial capacity~\cite{mcphee2010importance,conde2011emerging}; and how does captivity affect the ability of animals to be potentially reintroduced into the wild~\cite{kleiman2010wild}? Answering such questions via prolonged monitoring is particularly relevant for Great Apes where many gorilla species are critically endangered~\cite{walsh2003catastrophic}. Manual monitoring by specialists~\cite{jarvis2007effects}, however, is labour intensive.

\begin{figure}[t]
  \centering
  \includegraphics[width=\linewidth]{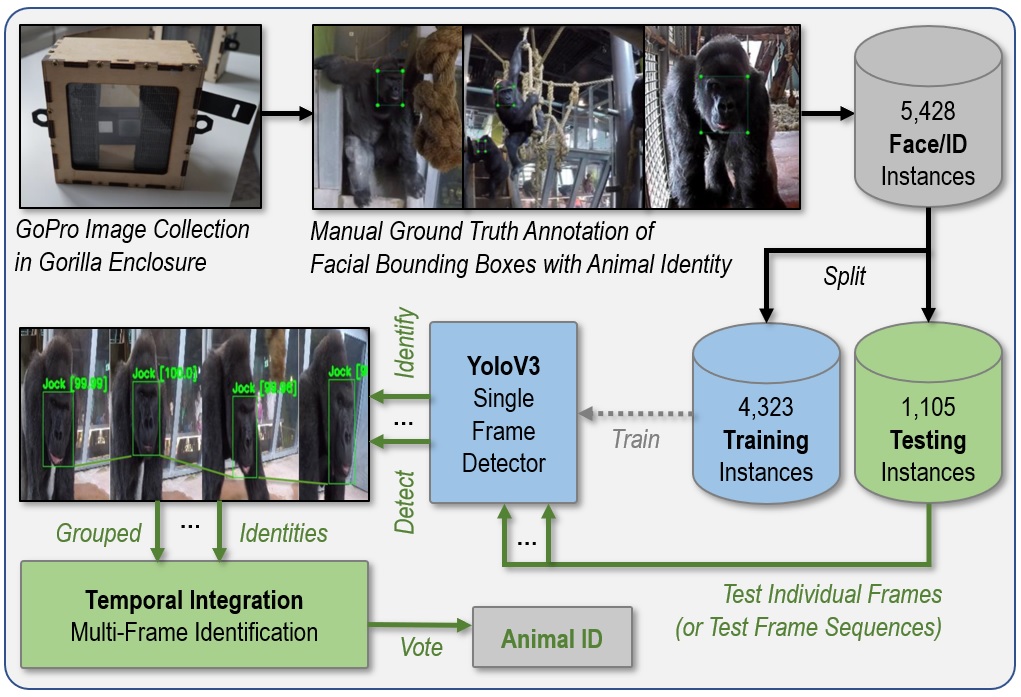}\vspace{-10pt}
  \caption{Data Collection, Training \& System Overview. \textmd{(a)~Ruggedly enclosed GoPro cameras were placed in a zoo enclosure housing a group of 7~lowland gorillas at Bristol Zoo Gardens. (b)~5k+~frames from the gathered video data were then manually annotated with a bounding box and identity information. (c)~This dataset was used to train and evaluate a YOLOv3 detection and classification framework, which was tested as (d)~a single frame recognition system, and a multi-frame system yielding (e)~location and identity data of the gorilla faces in view.}}
  \vspace{-8pt}
  \label{fig:overview}
\end{figure}

\textit{Contribution.} This paper provides a new annotated dataset for Great Ape facial ID and investigates how far YOLOv3~\cite{redmon2018yolov3} can be used to simultaneously detect and identify individual zoo gorillas based on facial characteristics (see Fig.~\ref{fig:overview}). Our contributions are: (1)~Collection and annotation of 5,428~samples of 7~western lowland gorillas~(see Fig.~\ref{fig:troop}); (2)~Training and evaluation of the YOLOv3 framework for single frame gorilla face localisation and classification, and (3)~Implementation of an offline multi-frame video application that delivers robust IDs in a zoo environment.

\section{Background \& Related Work}
\textit{Human Facial Biometrics.} Facial recognition technology for humans has long been a key field within computer vision~\cite{turk1991face, tolba2006face}. 
Deep convolutional neural networks~(CNNs)~\cite{hinton2015deep} exemplified in frameworks such as DeepFace~\cite{taigman2014deepface} form the base of most modern facial biometric frameworks~\cite{parkhi2015deep}.

\begin{figure}[!t]
  \centering
  \includegraphics[width=\linewidth]{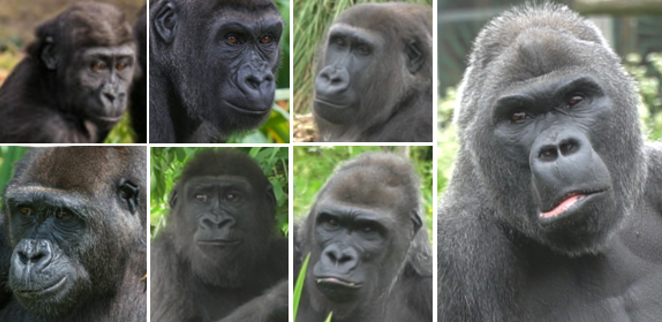}\vspace{-10pt}
  \caption{Lowland Gorilla Troop. \textmd{The figure depicts row-by-row left-to-right: Ayana, Kukuena, Kala, Touni, Afia, Kera. The large image on the far right is of Jock. The troop consists of 1 male and 6 females aged between 6 months and 37 years.}}
  \vspace{-8pt}
  \label{fig:troop}
\end{figure}

\textit{Great Ape Face Recognition.} Great Apes show facial characteristics that are individually distinct~(see Fig.\ref{fig:troop}) and not dissimilar to those of humans owing to our close evolutionary lineage~\cite{loos2011identification}. Thus, methodologies in animal biometrics~\cite{kuehl2013} follow approaches for human face recognition closely. Loos \textit{et al.}~\cite{loos2013automated} developed one of the first chimpanzee facial recognition systems based on traditional machine learning techniques. He also annotated key datasets~\cite{loos2013automated}, where further datasets have since been presented by Brust \textit{et al.}~\cite{brust2017towards} and Schofield \textit{et al.}~\cite{schofield2019chimpanzee}. Building on Loos's work and data, Freytag \textit{et al.}~\cite{freytag2016chimpanzee} trained a deep learning object detector, YOLOv2~\cite{redmon2017yolo9000}, to localise the faces of chimpanzees. They utilised a second deep CNN for feature extraction (AlexNet~\cite{krizhevsky2012imagenet} and VGGFaces~\cite{parkhi2015deep}), and a linear support vector machine (SVM)~\cite{svm2000} for identification. Later, Brust \textit{et al.}~\cite{brust2017towards} extended their work utilising a much larger and diverse dataset. Most recently, Schofield \textit{et al.}~\cite{schofield2019chimpanzee} presented a pipeline for identification of 23 chimpanzees across a video archive spanning 14 years. Similar to Brust~\textit{et al.}~\cite{brust2017towards}, a single-shot object detector, SSD~\cite{schofield2019chimpanzee}, is trained to perform localisation, and a secondary CNN model is trained to perform individual classification. They group video detections into tracklets across which identities are computed from, showing an improvement over single frame operation. In contrast to all discussed systems, in this paper we employ YOLOv3~\cite{redmon2018yolov3} to perform \textit{one-step} simultaneous facial detection and individual identification on gorillas, leading to a simpler yet equally effective pipeline.
Details of the existing great ape facial recognition systems and datasets can be found in the following literature; Loos \textit{et al.} 2011~\cite{loos2011identification}, Loos \textit{et al.} 2013~\cite{loos2013automated}, Freytag \textit{et al.} 2016~\cite{freytag2016chimpanzee}, Brust \textit{et al.} 2017~\cite{brust2017towards} and Schofield \textit{et al.} 2019~\cite{schofield2019chimpanzee}

\section{Dataset}
\textit{BristolGorillas2020.} The BristolGorillas2020 dataset comprises 628 video segments and 5,428 annotated facial images (sampled from the corresponding video segments). The train-test splits for the images were generated using stratified sampling (see Table~\ref{tab:data}). The test set for the single-frame and multi-frame detector are the same; each tracklet comprises detections made on the $n$ frames preceding the ground-truth frames included the test set. Data used in training is not evaluated during testing of the video application. The dataset GoPro (versions 5 \&\ 7) and Crosstour Action cameras~(see Fig.~\ref{fig:overview}) were fitted in the enclosure near enrichment devices~\cite{gray2018gorilla} to obtain close-up facial footage of the gorillas~(see Fig.~\ref{fig:annotation}). Data collection took place twice per week (from 11am to 1pm) over 6 weeks to record RGB video at 1280$\times$720 pixels and 30fps. A selection of frames containing front facial images was manually labelled~(see Fig.~\ref{fig:annotation}, top) with the help of experts from the Primate division at Bristol Zoo to ensure the identities of individual gorillas were labelled correctly.

\begin{table}[!h]
  \caption{Complete Dataset. \textmd{The table shows the total number of collected facial image patches, including the number of training and testing patches for each individual gorilla.}}
  \vspace{-5pt}
  \footnotesize{
  \begin{tabular}{|c|c|c|c|}
    \hline
    Individual Gorillas & Training & Testing & Total Images\\
    \hline
    Afia & 614 & 157 & 771\\
    Ayana & 489 & 126 & 615\\
    Jock & 387 & 101 & 488\\
    Kala & 578 & 148 & 726\\
    Kera & 776 & 196 & 972\\
    Kukuena & 747 & 190 & 937\\
    Touni & 732 & 187 & 919\\
    \hline
    \textbf{Total} & \textbf{4,323} & \textbf{1,105} & \textbf{5,428}\\
    \hline 
  \end{tabular}} \vspace{-8pt}
  \label{tab:data}
\end{table}

\section{Implementation}
\begin{figure}[!t]
  \centering
  \includegraphics[width=\linewidth]{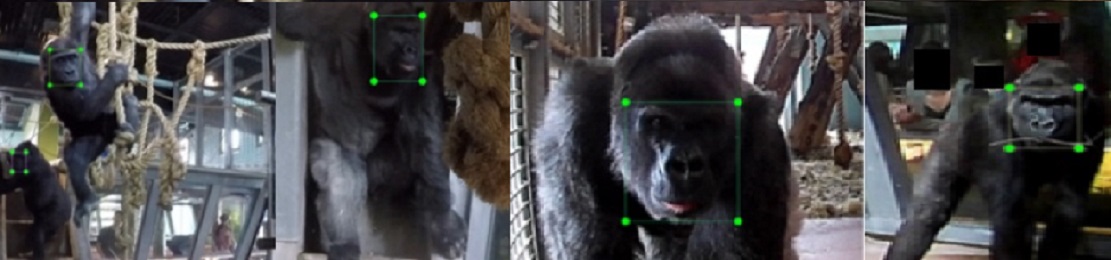}
  \vspace{-20pt}
  \caption{Gorilla Face Annotation. \textmd{ A collection of images annotated with front facial bounding boxes.}}
  \vspace{-12pt}
  \label{fig:annotation}
\end{figure}

\textit{Single-Frame Identification.} To train YOLOv3 we employed the open-source and freely available Darknet implementation pre-trained on ImageNet1000~\cite{redmon2018yolov3}. This network was then fine-tuned using stochastic gradient descent with momentum and a batch size of 32 at an input resolution of 416$\times$416 RGB pixels. Fine-tuning was performed with batch normalisation, data augmentation~(see Fig.~\ref{fig:annotation}, bottom), and learning rate decay (an initial learning rate of 0.001 reduced by a factor of 10 at 80\% and 90\% of the total training iterations). We trained against a one-hot encoded class vector, where each vector element represented a gorilla identity. The resulting YOLOv3 network forms the backbone of the facial recognition system by performing one-step multi-instance localisation and identification of gorilla faces.

\begin{figure}[H]
 \centering
  \includegraphics[width=\linewidth,height=70pt]{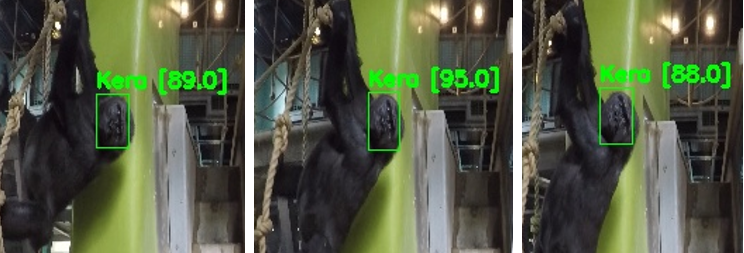}
  \vspace{-18pt}
  \caption{Multi-Frame Recognition. \textmd{The figure exemplifies a tracklet and shows confidence scores for the dominant class 'Kera' on a sequence where she swings across from a nest.}}
  \vspace{-10pt}
  \label{fig:tracklet}
\end{figure}

\textit{Multi-Frame Identification.} In Phase 2 the network is applied to individual frames of a sequence (i.e. tracklets across video) where \(X_t\) denotes the frame at time step~\textit{t}. All detections in~\(X_t\) and \(X_{t+1}\) are then input into a simple algorithm that uniquely associates cross-frame detections if they show the highest pairwise intersection-over-union (IoU) and this IoU is also greater than a threshold~$\theta=0.5$. The resulting association chains represent tracklets (see Fig.~\ref{fig:tracklet}). Their length ranges from a single frame to 10 frames. For each tracklet we evaluate identity classification via two methods: (1) highest single class probability denoted as \textit{maximum}, or the highest average class probability denoted as \textit{average}. For each detection we use the IoU assignment to compare the ground truth against the identity assigned to the tracklet containing the detection (where non-detections contribute to false negatives).

\section{Results}
\begin{figure}[H]
\centering
\includegraphics[width=\linewidth]{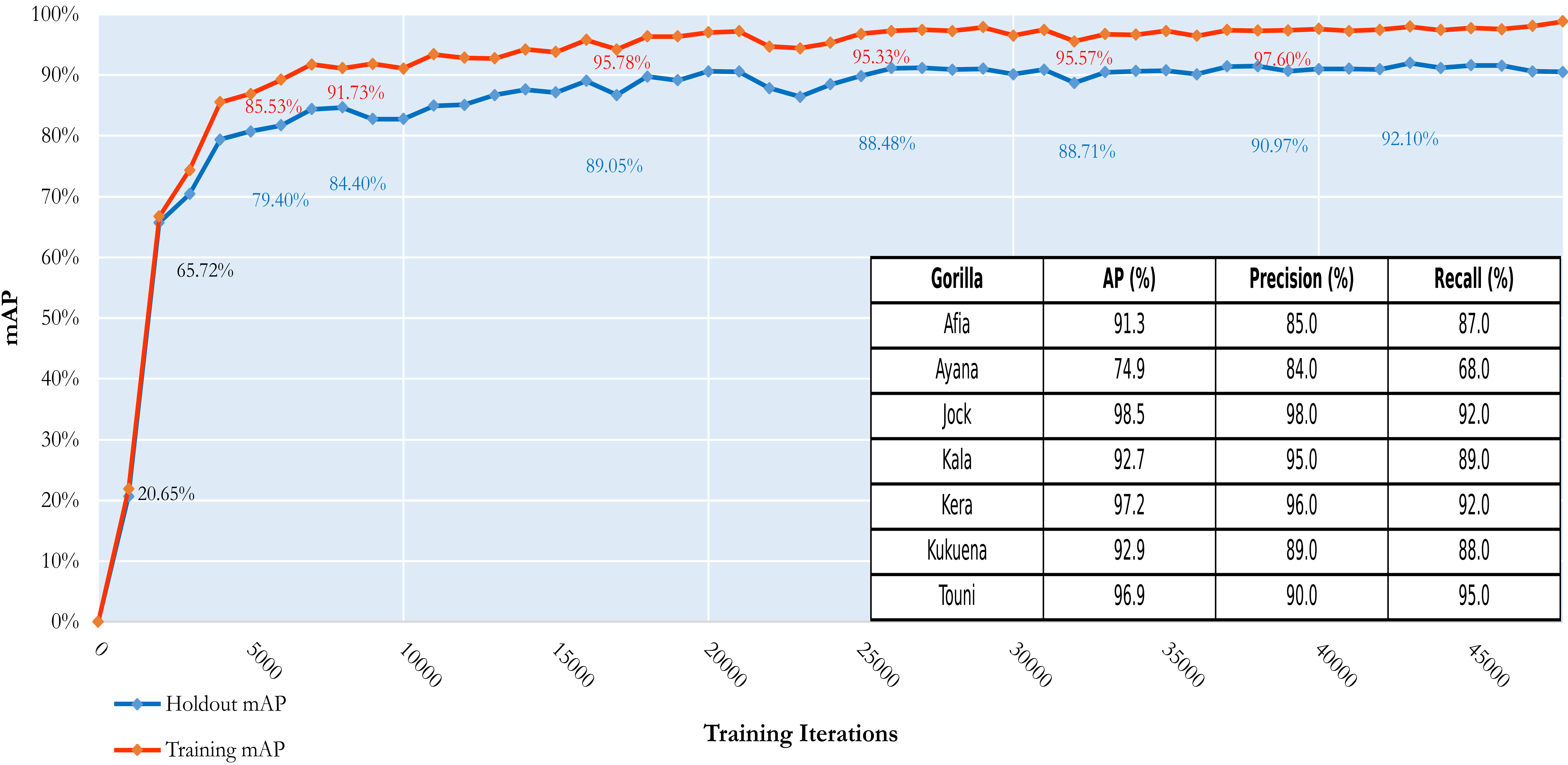}
\vspace{-18pt}
\caption{Single Frame YOLOv3 Identification Performance. \textmd{The graph shows the mAP scores on training~(red) and testing~(blue) sets against training iterations. The table depicts testing results for each of the individuals.}}
\vspace{-8pt}
\label{fig:perform}
\end{figure}

\begin{figure}[!t]
  \centering
  \includegraphics[width=\linewidth,height=70px]{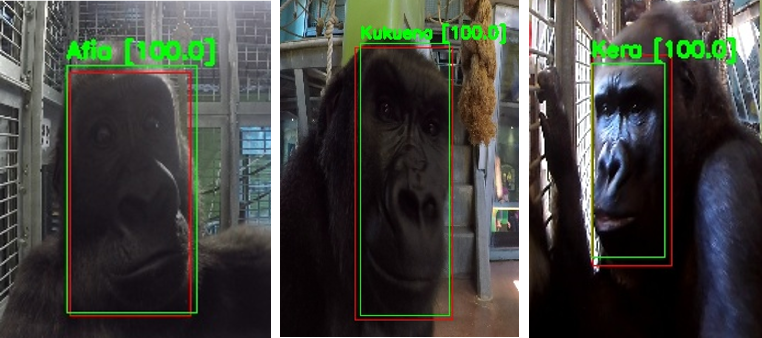}
  \vspace{-18pt}
  \caption{Localisation. \textmd{The figure shows examples of ground-truth bounding boxes~(red) and predicted bounding boxes~(green) for Afia (left), Kukuena (middle) and Kera (right).}}
  \vspace{-10pt}
  \label{fig:loc}
\end{figure}

\textit{Single Frame Detector.} Figure~\ref{fig:perform} reports single frame classification performance of YOLOv3 as precision, recall, and mean average precision (mAP) for the test set as well as mAP curves across the training process over 44k steps for both training~(red) and test~(blue) sets. We noted that the average IoU of this network was 77.92\%~(see Fig.~\ref{fig:loc}), whilst the average IoU scores for Afia and Ayana are only 71\% and 67\%, i.e. 6.92\% and 10.92\% below the network average, respectively. These are younger members of the troop; they are physically smaller in size and are often in the background rather than directly in front of the camera. This clearly challenged the network with regard to precise localisation and identification. Despite YOLOv3's improved detection capabilities~\cite{redmon2018yolov3}, ID classification suffers if image patches are too small. For instance,  Ayana’s AP score at 74.9\% is particularly low which may in part to be due to the fact that 64.34\% of her ground-truth annotations are smaller than $32\times 32$~pixels.

\textit{Multi-Frame Detector.} Table~\ref{tab:perform} reports multi-frame classification performance via precision, recall, and AP for the test set, where network weights are used which achieved the highest mAP score in single frame detection. The results reported utilise voting across a maximum tracklet size of 5, a stride of 1 and an IoU association threshold of 0.5. The multi-frame detector with maximum voting achieves the highest mAP, however, there is only a marginal difference between the maximum and average voting algorithms with less than 0.5\% difference between all three of the reported evaluation metrics. Both multi-frame detection approaches outperform the single frame detector across all metrics. The mAP improvements achieved by the average and maximum voting algorithms when compared with the single-frame detector are 5.2\% and 5.4\%, respectively.

\textit{Cross Validation.} We perform stratified 5-fold cross-validation on both single-frame and multi-frame identification systems. We train each fold for 24,000 iterations owing to time and computational restrictions. The three identification systems, single-frame and multi-frame identification with average and maximum voting schemes, achieve 89.91\%, 95.94\% and 96.65\% mAP, respectively.

\begin{table}[H]
\caption{Multi-Frame Detector Performance. \textmd{Performance is shown for both average and maximum voting schemes. The performance of the single-frame detector is included for comparison.}}
\vspace{-5pt}
\footnotesize{
\begin{tabular}{|c|c|c|c|}
\hline
Detection &mAP (\%)&Precision (\%)&Recall (\%)\\
\hline
Single & 92.1 (± 8.0) & 91.0 (± 5.5) & 87.3 (± 9.9)\\
Average & 97.3 (± 2.5) & 95.1 (± 4.7) & 91.1 (± 6.5)\\
Maximum & 97.5 (± 2.2) & 95.4 (± 2.7) & 91.2 (± 7.9)\\
 \hline
  \end{tabular}} \vspace{-8pt}
  \label{tab:perform}
\end{table}

\section{Ablation Studies}
\textit{Ablation Experiments.} We now investigate the effect of tracklet size, stride and association threshold on mAP performance. Produced based on evaluation on the test set the results are shown in Table~\ref{tab:ablation}. All three parameters prove stable with respect to changes. This indicates that the multi-frame identification improvement observed for the task at hand essentially rests with sampling multiple frames irrespective of the details of sampling.  

\begin{table}[!ht]
\caption{Ablation Experiments for Multi-Frame Detector. \textmd{The table presents results produced using the multi-frame detection approach with maximum and average voting schemes varying maximum tracklet length, temporal tracklet stride, and IoU association threshold for generating tracklets, respectively.}}
\begin{center}
\vspace{-5pt}
\footnotesize{
\begin{tabular}{| c | c | c | c |}
\cline{2-4}
\multicolumn{1}{c |}{}
& \multicolumn{3}{ c |}{mAP (\%)}\\ 
\hline\hline
Tracklet Length & 3 frames & 5 frames & 10 frames \\
\hline

Average & 97.2 (± 2.4) & 97.3 (± 2.5) & 97.4 (± 2.5) \\ \hline
Maximum & 97.4 (± 2.9) & 97.5 (± 2.2) & 97.5 (± 2.9) \\ \hline\hline
Tracklet Stride & 1 frame & 3 frames & 5 frames \\
\hline

Average & 97.3 (± 2.5) & 96.4 (± 3.4) & 96.5 (± 3.3) \\ \hline
Maximum & 97.5 (± 2.2) & 97.4 (± 3.4) & 97.5 (± 3.3) \\ \hline\hline
Association Threshold & IoU>0.25 & IoU>0.5 & IoU>0.75 \\
\hline
Average & 97.1 (± 2.7) & 97.3 (± 2.5) & 96.7 (± 2.6) \\ \hline
Maximum & 97.5 (± 2.9) & 97.5 (± 2.2) & 97.5 (± 2.7) \\ \hline
\end{tabular}}
\end{center}\vspace{-8pt}
\label{tab:ablation}
\end{table}

\section{Qualitative Discussion}
\textit{Detection Errors.} Most of the observed false-positives are attributable to ground-truths with a small bounding box (less than 32 x 32-pixel resolutions). The majority relate to either Afia or Ayana, the youngest members of the troop, who rarely appear in the forefront of footage and have the largest proportion of small ground truth bounding boxes. This suggests that YOLOv3 is less effective at detecting small objects although there are many other factors to consider. However, most of the remaining false-positive detections appear to be caused by extreme variations in pose.

\section{Conclusion}
\textit{Summary.} We presented a new dataset for the facial identification of Gorillas~(\textit{Gorilla gorilla gorilla}) which contains 5k+ annotated frames. We evaluated a multi-frame deep learning system based on the YOLOv3 framework on the task of facially recognising 7~individual western lowland gorillas in a zoo environment. We showed that the system is able to perform identifications above~92\%~mAP when operating on single frames and above~97\%~mAP when operating on tracklets. We conclude that, despite the availability of more complex systems, the proposed straight forward end-to-end application as presented operates sufficiently robustly to be used for enriching the research capabilities of zoo environments as well as their visitor experience.

\textit{Future Work.} We intend to train our system on the dataset compiled by Brust et al \cite{brust2017towards} and Schofield et al \cite{schofield2019chimpanzee} for comparison. Furthermore, preparatory work to install our system as a permanent feature at Bristol Zoo is underway. Footage from a camera fitted in the enclosure will be streamed to a screen in the visitor area and display the identities of individual gorillas. It is hoped that this will improve the visitor experience and help to raise revenue for the zoo to invest in further research or conservation programs.

\section{Acknowledgements}

We would like to acknowledge the contributions of our collaborators for their help in facilitating the technical innovation described in this paper. Specifically, we’d like to thank the other members of the Gorilla Game Lab team, Dr Fay Clark, Dr Katy Burgess, Dr Peter Bennett, and Dr Stuart Gray for their efforts. We would also like to extend our thanks to the staff at Bristol Zoo Gardens, most notably to the keepers, and to Betty Roberts.

\bibliographystyle{ACM-Reference-Format}
\footnotesize{

\begin{thebibliography}{22}


\ifx \showCODEN    \undefined \def \showCODEN     #1{\unskip}     \fi
\ifx \showDOI      \undefined \def \showDOI       #1{#1}\fi
\ifx \showISBNx    \undefined \def \showISBNx     #1{\unskip}     \fi
\ifx \showISBNxiii \undefined \def \showISBNxiii  #1{\unskip}     \fi
\ifx \showISSN     \undefined \def \showISSN      #1{\unskip}     \fi
\ifx \showLCCN     \undefined \def \showLCCN      #1{\unskip}     \fi
\ifx \shownote     \undefined \def \shownote      #1{#1}          \fi
\ifx \showarticletitle \undefined \def \showarticletitle #1{#1}   \fi
\ifx \showURL      \undefined \def \showURL       {\relax}        \fi
\providecommand\bibfield[2]{#2}
\providecommand\bibinfo[2]{#2}
\providecommand\natexlab[1]{#1}
\providecommand\showeprint[2][]{arXiv:#2}

\bibitem[\protect\citeauthoryear{Brust, Burghardt, Groenenberg, Kading, Kuhl,
  Manguette, and Denzler}{Brust et~al\mbox{.}}{2017}]%
        {brust2017towards}
\bibfield{author}{\bibinfo{person}{Clemens-Alexander Brust},
  \bibinfo{person}{Tilo Burghardt}, \bibinfo{person}{Milou Groenenberg},
  \bibinfo{person}{Christoph Kading}, \bibinfo{person}{Hjalmar~S Kuhl},
  \bibinfo{person}{Marie~L Manguette}, {and} \bibinfo{person}{Joachim
  Denzler}.} \bibinfo{year}{2017}\natexlab{}.
\newblock \showarticletitle{Towards automated visual monitoring of individual
  gorillas in the wild}. In \bibinfo{booktitle}{\emph{Proceedings of the IEEE
  International Conference on Computer Vision}}. \bibinfo{pages}{2820--2830}.
\newblock


\bibitem[\protect\citeauthoryear{Conde, Flesness, Colchero, Jones, Scheuerlein,
  et~al\mbox{.}}{Conde et~al\mbox{.}}{2011}]%
        {conde2011emerging}
\bibfield{author}{\bibinfo{person}{Dalia~Amor Conde}, \bibinfo{person}{Nate
  Flesness}, \bibinfo{person}{Fernando Colchero}, \bibinfo{person}{Owen~R
  Jones}, \bibinfo{person}{Alexander Scheuerlein}, {et~al\mbox{.}}}
  \bibinfo{year}{2011}\natexlab{}.
\newblock \showarticletitle{An emerging role of zoos to conserve biodiversity}.
\newblock \bibinfo{journal}{\emph{Science}} \bibinfo{volume}{331},
  \bibinfo{number}{6023} (\bibinfo{year}{2011}), \bibinfo{pages}{1390--1391}.
\newblock


\bibitem[\protect\citeauthoryear{Cristianini, Shawe-Taylor,
  et~al\mbox{.}}{Cristianini et~al\mbox{.}}{2000}]%
        {svm2000}
\bibfield{author}{\bibinfo{person}{Nello Cristianini}, \bibinfo{person}{John
  Shawe-Taylor}, {et~al\mbox{.}}} \bibinfo{year}{2000}\natexlab{}.
\newblock \bibinfo{booktitle}{\emph{An introduction to support vector machines
  and other kernel-based learning methods}}.
\newblock \bibinfo{publisher}{Cambridge university press}.
\newblock


\bibitem[\protect\citeauthoryear{Fraser}{Fraser}{2009}]%
        {fraser2009assessing}
\bibfield{author}{\bibinfo{person}{David Fraser}.}
  \bibinfo{year}{2009}\natexlab{}.
\newblock \showarticletitle{Assessing animal welfare: different philosophies,
  different scientific approaches}.
\newblock \bibinfo{journal}{\emph{Zoo Biology: Published in affiliation with
  the American Zoo and Aquarium Association}} \bibinfo{volume}{28},
  \bibinfo{number}{6} (\bibinfo{year}{2009}), \bibinfo{pages}{507--518}.
\newblock


\bibitem[\protect\citeauthoryear{Freytag, Rodner, Simon, Loos, K{\"u}hl, and
  Denzler}{Freytag et~al\mbox{.}}{2016}]%
        {freytag2016chimpanzee}
\bibfield{author}{\bibinfo{person}{Alexander Freytag}, \bibinfo{person}{Erik
  Rodner}, \bibinfo{person}{Marcel Simon}, \bibinfo{person}{Alexander Loos},
  \bibinfo{person}{Hjalmar~S K{\"u}hl}, {and} \bibinfo{person}{Joachim
  Denzler}.} \bibinfo{year}{2016}\natexlab{}.
\newblock \showarticletitle{Chimpanzee faces in the wild: Log-euclidean cnns
  for predicting identities and attributes of primates}. In
  \bibinfo{booktitle}{\emph{German Conference on Pattern Recognition}}.
  Springer, \bibinfo{pages}{51--63}.
\newblock


\bibitem[\protect\citeauthoryear{Gray, Clark, Burgess, Metcalfe, Kadijevic,
  Cater, and Bennett}{Gray et~al\mbox{.}}{2018}]%
        {gray2018gorilla}
\bibfield{author}{\bibinfo{person}{Stuart Gray}, \bibinfo{person}{Fay Clark},
  \bibinfo{person}{Katy Burgess}, \bibinfo{person}{Tom Metcalfe},
  \bibinfo{person}{Anja Kadijevic}, \bibinfo{person}{Kirsten Cater}, {and}
  \bibinfo{person}{Peter Bennett}.} \bibinfo{year}{2018}\natexlab{}.
\newblock \showarticletitle{Gorilla Game Lab: Exploring modularity, tangibility
  and playful engagement in cognitive enrichment design}. In
  \bibinfo{booktitle}{\emph{Proceedings of the Fifth International Conference
  on Animal-Computer Interaction}}. \bibinfo{pages}{1--13}.
\newblock


\bibitem[\protect\citeauthoryear{Jarvis}{Jarvis}{2007}]%
        {jarvis2007effects}
\bibfield{author}{\bibinfo{person}{Kiersten~Austad Jarvis}.}
  \bibinfo{year}{2007}\natexlab{}.
\newblock \emph{\bibinfo{title}{Effects Of a Complex Enrichment Device On Tool
  Use, Tool Manufacturing, Activity Budgets, And Stereotypic Behaviors In
  Captive Western Lowland Gorillas (Gorilla gorilla gorilla)}}.
\newblock \bibinfo{thesistype}{Ph.D. Dissertation}. \bibinfo{school}{University
  of West Florida}.
\newblock


\bibitem[\protect\citeauthoryear{Kleiman, Thompson, and Baer}{Kleiman
  et~al\mbox{.}}{2010}]%
        {kleiman2010wild}
\bibfield{author}{\bibinfo{person}{Devra~G Kleiman},
  \bibinfo{person}{Katerina~V Thompson}, {and} \bibinfo{person}{Charlotte~Kirk
  Baer}.} \bibinfo{year}{2010}\natexlab{}.
\newblock \bibinfo{booktitle}{\emph{Wild mammals in captivity: principles and
  techniques for zoo management}}.
\newblock \bibinfo{publisher}{University of Chicago Press}.
\newblock


\bibitem[\protect\citeauthoryear{Krizhevsky, Sutskever, and Hinton}{Krizhevsky
  et~al\mbox{.}}{2012}]%
        {krizhevsky2012imagenet}
\bibfield{author}{\bibinfo{person}{Alex Krizhevsky}, \bibinfo{person}{Ilya
  Sutskever}, {and} \bibinfo{person}{Geoffrey~E Hinton}.}
  \bibinfo{year}{2012}\natexlab{}.
\newblock \showarticletitle{Imagenet classification with deep convolutional
  neural networks}. In \bibinfo{booktitle}{\emph{Advances in neural information
  processing systems}}. \bibinfo{pages}{1097--1105}.
\newblock


\bibitem[\protect\citeauthoryear{Kuehl and Burghardt}{Kuehl and
  Burghardt}{2013}]%
        {kuehl2013}
\bibfield{author}{\bibinfo{person}{Hjalmar Kuehl} {and} \bibinfo{person}{Tilo
  Burghardt}.} \bibinfo{year}{2013}\natexlab{}.
\newblock \showarticletitle{Animal biometrics: Quantifying and detecting
  phenotypic appearance}.
\newblock \bibinfo{journal}{\emph{Trends in ecology \& evolution}}
  \bibinfo{volume}{28} (\bibinfo{date}{03} \bibinfo{year}{2013}).
\newblock
\urldef\tempurl%
\url{https://doi.org/10.1016/j.tree.2013.02.013}
\showDOI{\tempurl}


\bibitem[\protect\citeauthoryear{LeCun, Bengio, and Hinton}{LeCun
  et~al\mbox{.}}{2015}]%
        {hinton2015deep}
\bibfield{author}{\bibinfo{person}{Yann LeCun}, \bibinfo{person}{Yoshua
  Bengio}, {and} \bibinfo{person}{Geoffrey Hinton}.}
  \bibinfo{year}{2015}\natexlab{}.
\newblock \showarticletitle{Deep learning}.
\newblock \bibinfo{journal}{\emph{nature}} \bibinfo{volume}{521},
  \bibinfo{number}{7553} (\bibinfo{year}{2015}), \bibinfo{pages}{436--444}.
\newblock


\bibitem[\protect\citeauthoryear{Loos and Ernst}{Loos and Ernst}{2013}]%
        {loos2013automated}
\bibfield{author}{\bibinfo{person}{Alexander Loos} {and}
  \bibinfo{person}{Andreas Ernst}.} \bibinfo{year}{2013}\natexlab{}.
\newblock \showarticletitle{An automated chimpanzee identification system using
  face detection and recognition}.
\newblock \bibinfo{journal}{\emph{EURASIP Journal on Image and Video
  Processing}} \bibinfo{volume}{2013}, \bibinfo{number}{1}
  (\bibinfo{year}{2013}), \bibinfo{pages}{49}.
\newblock


\bibitem[\protect\citeauthoryear{Loos, Pfitzer, and Aporius}{Loos
  et~al\mbox{.}}{2011}]%
        {loos2011identification}
\bibfield{author}{\bibinfo{person}{Alexander Loos}, \bibinfo{person}{Martin
  Pfitzer}, {and} \bibinfo{person}{Laura Aporius}.}
  \bibinfo{year}{2011}\natexlab{}.
\newblock \showarticletitle{Identification of great apes using face
  recognition}. In \bibinfo{booktitle}{\emph{2011 19th European Signal
  Processing Conference}}. IEEE, \bibinfo{pages}{922--926}.
\newblock


\bibitem[\protect\citeauthoryear{McPhee and Carlstead}{McPhee and
  Carlstead}{2010}]%
        {mcphee2010importance}
\bibfield{author}{\bibinfo{person}{M~Elsbeth McPhee} {and}
  \bibinfo{person}{Kathy Carlstead}.} \bibinfo{year}{2010}\natexlab{}.
\newblock \showarticletitle{The importance of maintaining natural behaviors in
  captive mammals}.
\newblock \bibinfo{journal}{\emph{Wild mammals in captivity: Principles and
  techniques for zoo management}}  \bibinfo{volume}{2} (\bibinfo{year}{2010}),
  \bibinfo{pages}{303--313}.
\newblock


\bibitem[\protect\citeauthoryear{Parkhi, Vedaldi, Zisserman,
  et~al\mbox{.}}{Parkhi et~al\mbox{.}}{2015}]%
        {parkhi2015deep}
\bibfield{author}{\bibinfo{person}{Omkar~M Parkhi}, \bibinfo{person}{Andrea
  Vedaldi}, \bibinfo{person}{Andrew Zisserman}, {et~al\mbox{.}}}
  \bibinfo{year}{2015}\natexlab{}.
\newblock \showarticletitle{Deep face recognition.}. In
  \bibinfo{booktitle}{\emph{bmvc}}, Vol.~\bibinfo{volume}{1}.
  \bibinfo{pages}{6}.
\newblock


\bibitem[\protect\citeauthoryear{Redmon and Farhadi}{Redmon and
  Farhadi}{2017}]%
        {redmon2017yolo9000}
\bibfield{author}{\bibinfo{person}{Joseph Redmon} {and} \bibinfo{person}{Ali
  Farhadi}.} \bibinfo{year}{2017}\natexlab{}.
\newblock \showarticletitle{YOLO9000: better, faster, stronger}. In
  \bibinfo{booktitle}{\emph{Proceedings of the IEEE conference on computer
  vision and pattern recognition}}. \bibinfo{pages}{7263--7271}.
\newblock


\bibitem[\protect\citeauthoryear{Redmon and Farhadi}{Redmon and
  Farhadi}{2018}]%
        {redmon2018yolov3}
\bibfield{author}{\bibinfo{person}{Joseph Redmon} {and} \bibinfo{person}{Ali
  Farhadi}.} \bibinfo{year}{2018}\natexlab{}.
\newblock \showarticletitle{Yolov3: An incremental improvement}.
\newblock \bibinfo{journal}{\emph{arXiv preprint arXiv:1804.02767}}
  (\bibinfo{year}{2018}).
\newblock


\bibitem[\protect\citeauthoryear{Schofield, Nagrani, Zisserman, Hayashi,
  Matsuzawa, Biro, and Carvalho}{Schofield et~al\mbox{.}}{2019}]%
        {schofield2019chimpanzee}
\bibfield{author}{\bibinfo{person}{Daniel Schofield}, \bibinfo{person}{Arsha
  Nagrani}, \bibinfo{person}{Andrew Zisserman}, \bibinfo{person}{Misato
  Hayashi}, \bibinfo{person}{Tetsuro Matsuzawa}, \bibinfo{person}{Dora Biro},
  {and} \bibinfo{person}{Susana Carvalho}.} \bibinfo{year}{2019}\natexlab{}.
\newblock \showarticletitle{Chimpanzee face recognition from videos in the wild
  using deep learning}.
\newblock \bibinfo{journal}{\emph{Science Advances}} \bibinfo{volume}{5},
  \bibinfo{number}{9} (\bibinfo{year}{2019}), \bibinfo{pages}{eaaw0736}.
\newblock


\bibitem[\protect\citeauthoryear{Taigman, Yang, Ranzato, and Wolf}{Taigman
  et~al\mbox{.}}{2014}]%
        {taigman2014deepface}
\bibfield{author}{\bibinfo{person}{Yaniv Taigman}, \bibinfo{person}{Ming Yang},
  \bibinfo{person}{Marc'Aurelio Ranzato}, {and} \bibinfo{person}{Lior Wolf}.}
  \bibinfo{year}{2014}\natexlab{}.
\newblock \showarticletitle{Deepface: Closing the gap to human-level
  performance in face verification}. In \bibinfo{booktitle}{\emph{Proceedings
  of the IEEE conference on computer vision and pattern recognition}}.
  \bibinfo{pages}{1701--1708}.
\newblock


\bibitem[\protect\citeauthoryear{Tolba, El-Baz, and El-Harby}{Tolba
  et~al\mbox{.}}{2006}]%
        {tolba2006face}
\bibfield{author}{\bibinfo{person}{AS Tolba}, \bibinfo{person}{AH El-Baz},
  {and} \bibinfo{person}{AA El-Harby}.} \bibinfo{year}{2006}\natexlab{}.
\newblock \showarticletitle{Face recognition: A literature review}.
\newblock \bibinfo{journal}{\emph{International Journal of Signal Processing}}
  \bibinfo{volume}{2}, \bibinfo{number}{2} (\bibinfo{year}{2006}),
  \bibinfo{pages}{88--103}.
\newblock


\bibitem[\protect\citeauthoryear{Turk and Pentland}{Turk and Pentland}{1991}]%
        {turk1991face}
\bibfield{author}{\bibinfo{person}{Matthew~A Turk} {and}
  \bibinfo{person}{Alex~P Pentland}.} \bibinfo{year}{1991}\natexlab{}.
\newblock \showarticletitle{Face recognition using eigenfaces}. In
  \bibinfo{booktitle}{\emph{Proceedings. 1991 IEEE Computer Society Conference
  on Computer Vision and Pattern Recognition}}. IEEE,
  \bibinfo{pages}{586--591}.
\newblock


\bibitem[\protect\citeauthoryear{Walsh, Abernethy, Bermejo, Beyers, De~Wachter,
  Akou, Huijbregts, Mambounga, Toham, Kilbourn, et~al\mbox{.}}{Walsh
  et~al\mbox{.}}{2003}]%
        {walsh2003catastrophic}
\bibfield{author}{\bibinfo{person}{Peter~D Walsh}, \bibinfo{person}{Kate~A
  Abernethy}, \bibinfo{person}{Magdalena Bermejo}, \bibinfo{person}{Rene
  Beyers}, \bibinfo{person}{Pauwel De~Wachter}, \bibinfo{person}{Marc~Ella
  Akou}, \bibinfo{person}{Bas Huijbregts}, \bibinfo{person}{Daniel~Idiata
  Mambounga}, \bibinfo{person}{Andre~Kamdem Toham}, \bibinfo{person}{Annelisa~M
  Kilbourn}, {et~al\mbox{.}}} \bibinfo{year}{2003}\natexlab{}.
\newblock \showarticletitle{Catastrophic ape decline in western equatorial
  Africa}.
\newblock \bibinfo{journal}{\emph{Nature}} \bibinfo{volume}{422},
  \bibinfo{number}{6932} (\bibinfo{year}{2003}), \bibinfo{pages}{611}.
\newblock


\end{thebibliography}

}
\end{document}